\documentclass[a4paper,twoside]{article}

\usepackage[table,xcdraw]{xcolor}

\usepackage[ruled,vlined]{algorithm2e}
\usepackage{algpseudocode}
\usepackage{epsfig}
\usepackage{subcaption}
\usepackage{calc}
\usepackage{amssymb}
\usepackage{comment}
\usepackage{amstext}
\usepackage{amsmath}
\usepackage{booktabs}
\usepackage{amsthm}
\usepackage{svg}
\usepackage{multicol}
\usepackage{pslatex}
\usepackage{apalike}
\usepackage{hyperref}
\usepackage{apacite}

\usepackage[bottom]{footmisc}
\usepackage{SCITEPRESS}     

\begin{document}

\title{PDNet: Exploring Pruning and Distillation for Real-Time Worker Activity Recognition in Industrial Assembly}

\author{\authorname{Malek Baba\sup{1}\orcidAuthor{0009-0009-4927-6861}, Thinesh Thiyakesan Ponbagavathi\sup{2}\orcidAuthor{0009-0009-3920-1233} and Alina Roitberg
\sup{3}\orcidAuthor{0000-0003-4724-9164}}
\affiliation{\sup{1}Stuttgart University, Stuttgart, Germany}
\affiliation{\sup{2}Institute for Artificial Intelligence (KI), Stuttgart University, Stuttgart, Germany}
\affiliation{\sup{3}Intelligent Assistive Systems (IAS) group, University of Hildesheim, Hildesheim, Germany}
\email{Malek.baba@outlook.com, thinesh.thiyakesan-ponbagavathi@ki.uni-stuttgart.de, roitberg@uni-hildesheim.de}
}

\abstract{
In this work, we address real-time fine-grained human activity recognition (HAR) in industrial assembly and introduce \textsl{PDNet}, a  new framework that  jointly exploits structured pruning and knowledge distillation for 3D convolutional networks.
We start by training a 3D-CNN backbone in a supervised manner. Then, PDNet applies channel-wise structured pruning to obtain a more compact network. 
In the third step, our method is optimized using knowledge distillation from the large unpruned teacher to recover accuracy lost due to pruning.
On the challenging MECCANO dataset with an X3D-M backbone, the pruned model without distillation reaches a Top-1 accuracy of 25.54 \%, while PDNet raises this to 39.46 \%, with only 6.90 MB model size and 3.78 GFLOPs.
These results show that distillation within PDNet is highly effective in restoring pruned model performance and highlight its high potential  for resource-constrained HAR in industrial assembly.
}


\onecolumn \maketitle \normalsize \setcounter{footnote}{0} \vfill

\section{\uppercase{Introduction}}
In the context of modern manufacturing, often framed under the paradigm of Industry 4.0, artificial intelligence and, in particular, deep learning have become a key enabler.
An important element of such systems is the \textit{real-time analysis of worker activities}, which enables the quantification and evaluation of human performance during complex assembly operations~\cite {tao2018,chen2022}.
By accurately recognizing and predicting worker actions, these intelligent systems can provide immediate assistance and guidance, thereby optimizing assembly processes and ensuring adherence to standard operating procedures.
This capability is also essential for fostering effective human-robot collaboration \cite{almin2020} and can significantly reduce operational errors, directly advancing the core objectives of smart manufacturing~\cite{chen2020}.


\begin{figure}[!t]
    \centering
    \includegraphics[width=1\linewidth]{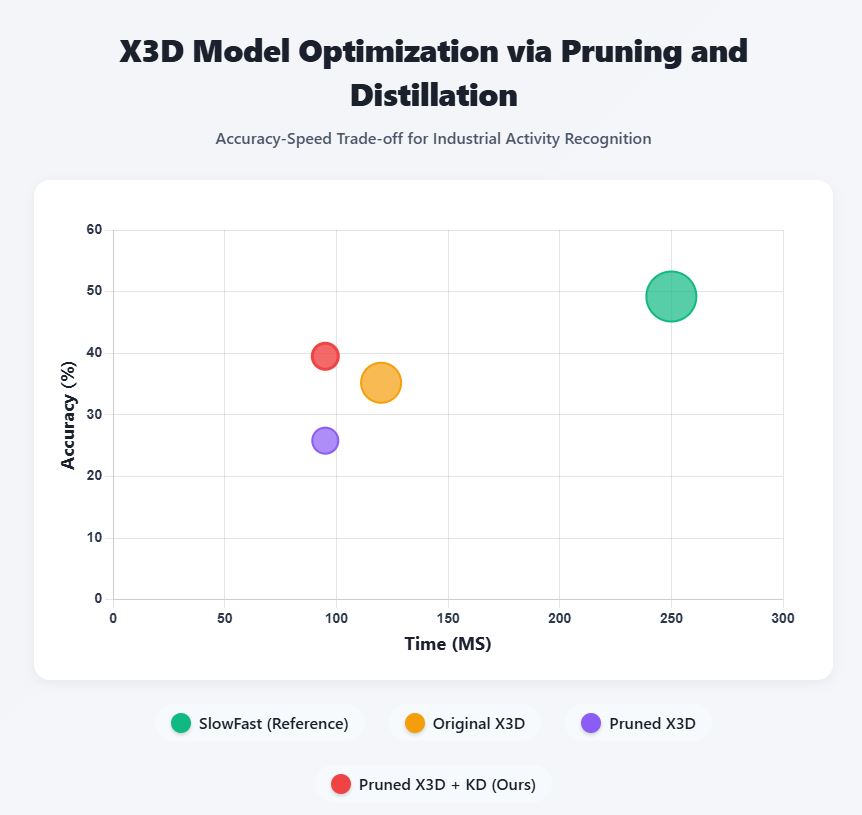}
    \caption{
    Accuracy-latency-model size trade-off for fine-grained industrial activity recognition on the MECCANO dataset. The X-axis shows CPU inference time per clip (ms) and the Y-axis Top-1 accuracy, the circle area encodes model size. Our pruned X3D with knowledge distillation (PDNet, red) recovers most of the accuracy lost by the pruned baseline (purple).
    }
    \label{fig:performance_analysis}
\end{figure}

Deep learning-based human activity recognition models, for example architectures based on 3D convolutional neural networks (3D CNNs)~\cite{3DCNNHAR}, play a key role in this context.
Such models can jointly capture spatial appearance and temporal dynamics from video streams, enabling accurate identification of complex actions \cite{3DCNNHAR}.
Their ability to model complex motion patterns makes them especially suitable for industrial settings, where accurately recognizing subtle worker activities is critical \cite{tran2015learningspatiotemporalfeatures3d}.

However, by extending convolutions into the temporal dimension, 3D architectures typically require substantially more parameters and computations than their 2D counterparts, which complicates deployment under tight resource constraints. Standard architectures such as I3D-ResNet50 exceed 25 GFLOPs, which makes them unsuitable for real-time or edge applications~\cite{Hedegaard_2022}. Efficient architectural designs, such as lightweight 3D backbones, aim to balance computational cost and accuracy~\cite{x3d}. Additional strategies, including model quantization and pruning, are often applied to reduce model size and inference latency, which supports deployment on resource-constrained embedded systems and FPGA accelerators~\cite{FMM-X3D}. Together, these techniques are crucial for achieving real-time operation in industrial environments~\cite{FMM-X3D}. In addition to architecture-level and compression strategies, input-level efficiency has also been explored. For example, Noor et al.~\cite{Noor2023ALS} propose a lightweight skeleton-based 3D-CNN that operates directly on sequences of human joint keypoints rather than RGB frames. By using pose sequences as input, the network drastically reduces input dimensionality and removes irrelevant background information, enabling a compact 3D-CNN to achieve real-time inference with low latency while maintaining strong accuracy for fall detection and general action recognition. At the same time, another compression technique, knowledge distillation, has gained traction across many domains and has been explored to improve accuracy of heavily compressed models for driver activity recognition~\cite{Tanama2023QuantizedDO}. Yet, there is little work that explores pruning and knowledge distillation for 3D CNNs in general activity recognition as well as in the industrial assembly domain: most existing efforts either target 2D architectures or apply compression techniques in isolation.

To close this gap, we introduce a new framework that jointly leverages structured pruning and knowledge distillation on top of a 3D-CNN for fine-grained recognition of industrial assembly activities, enabling model compression while effectively recovering the accuracy typically lost during pruning.
Overall, the key contributions of this work are as follows:
\\
\begin{itemize}
\item 
We first establish a strong baseline by benchmarking several compact 3D CNN architectures—including 3D-MobileNet~\cite{3dmobilenet} and X3D variants (X3D-M, X3D-L)~\cite{x3d}—on the MECCANO dataset for fine-grained industrial assembly recognition.  
This analysis compares accuracy, inference time, and model size under realistic deployment constraints and identifies \textbf{X3D-M} as a promising starting point for further optimization.

\item \textbf{Introducing Structured Pruning for 3D-CNN Efficiency:}  
To the best of our knowledge, we are the first to explore \textit{structured pruning} for efficiency optimization of 3D-CNNs in industrial assembly scenarios. 
Using L1-norm-based channel pruning at multiple sparsity levels, we significantly reduce computational cost and memory usage while quantifying the corresponding accuracy degradation.  
This step reveals both the potential and the limitations of pruning-only compression for fine-grained 3D video understanding.

\item \textbf{PDNet:  Pruning-Distillation Method for Fine-Grained Industrial HAR:}
Building on the pruning analysis, we introduce \textbf{PDNet}, a framework that couples L1-norm–based structured pruning of X3D-M with knowledge distillation from a high-capacity SlowFast teacher~\cite{slowfast}.
Within PDNet, pruning  reduces parameters and FLOPs, while distillation is explicitly used to recover the accuracy lost by compression, resulting in a compact 3D-CNN that maintains competitive performance for fine-grained recognition of industrial assembly activities under  resource constraints.

\end{itemize}

\section{Related Work}

Recent studies have employed a combination of lightweight model architectures, model compression techniques, and hardware optimization strategies to achieve improved performance. Lightweight 2D CNN-based networks such as MobileNet \cite{Prathima2025RealTimeIO} and EfficientNet \cite{Narayan2023EnhanceNetAA} are frequently adopted for their ability to maintain reasonable accuracy while significantly reducing computational load, making them suitable for deployment on edge devices. Model compression techniques are also widely applied; many existing approaches focus on reducing the memory footprint and latency of deep neural networks while preserving high classification accuracy. One key strategy involves model compression techniques, such as pruning and quantization. Structured pruning methods eliminate entire filters or channels, which align better with hardware execution patterns compared to unstructured sparsity \cite{he2023structured}. Techniques like DepGraph \cite{torchPruningLibrary} further extend structured pruning by incorporating graph representations of neural architectures, enabling fine-grained control over network topology. Quantization reduces model precision (e.g., from 32-bit floating-point to 8-bit integers), and post-training quantization (PTQ) and quantization-aware training (QAT) approaches have shown promising results for maintaining accuracy on industrial workloads \cite{Novac2021Quantization}. Knowledge distillation (KD) compresses large teacher models into lightweight student models without significant loss in accuracy \cite{hinton2015KD}, with recent work exploring its combination with quantization for additional efficiency gains \cite{Tanama2023QuantizedDO}.

\section{\uppercase{Methodology}}



We introduce \textbf{PDNet}, a unified optimization framework designed to obtain resource-efficient 3D-CNNs for fine-grained industrial activity recognition. 
PDNet consists of three components that operate sequentially.

First, we train a compact 3D convolutional neural network (3D-CNN) as our baseline model, described in Section~\ref{sec:baseline_training_placeholder}. 
This backbone provides the initial spatiotemporal feature extractor from which all subsequent optimizations are derived.
Second, we apply L1-norm-based \emph{structured pruning} (Section~\ref{sec:pruning_placeholder}) to reduce the parameter count and computational complexity of the baseline model. 
This step produces a compact student network with substantially lower resource demands but also introduces a measurable drop in recognition accuracy.
Finally, we incorporate \emph{knowledge distillation} (Section~\ref{sec:kd_placeholder}), using a high-capacity teacher model to guide the training of the pruned student. 
This distillation phase recovers much of the performance lost during pruning, resulting in an efficient 3D-CNN that maintains competitive accuracy under strict deployment constraints.

\subsection{PDNet: Pruning and Distillation for  Recognition of Assembly Activities}

\subsubsection{Backbone 3D-CNN}
\label{sec:baseline_training_placeholder}

First, we select an efficient 3D-CNN backbone. 
To this end, we benchmark several lightweight 3D architectures on the MECCANO dataset, focusing on the trade-off between accuracy, parameter count, and inference cost.
3D-MobileNet \cite{3dmobilenet} adapts the MobileNetV2 design by utilizing depthwise separable convolutions, which factorize standard 3D kernels to drastically reduce floating-point operations. Alternatively, the X3D family \cite{x3d} optimizes efficiency through a multi-axis expansion strategy, scaling dimensions such as frame rate, duration, and network width independently rather than uniformly. X3D offers a spectrum of variants ranging from the ultra-compact X3D-XS (Extra Small) and  X3D-S, designed for edge devices, to larger models like X3D-M, providing a superior trade-off between accuracy and computational complexity compared to traditional heavy models.
Based on this empirical comparison describt in the experiments section, we adopt \textit{X3D-M} as our baseline 3D-CNN. This model is first optimized in a supervised fashion using the standard cross-entropy loss, and then serves as the starting point for the structured pruning and knowledge distillation stages of PDNet.

\subsubsection{Structured Pruning}
\label{sec:pruning_placeholder}

The second stage of PDNet applies \emph{structured pruning} to reduce model complexity while preserving a hardware-friendly network structure \cite{torchPruningLibrary}. 
Structured pruning accelerates neural networks by removing groups of parameters such as channels or filters, while maintaining structural integrity for efficient inference~\cite{torchPruningLibrary}. Often, the importance of groups is measured by the L1-norm of their weights, which promotes sparsity by encouraging the sum of absolute values of weights to be small. 
\\
\\
A central concept is the pruning ratio \(p\), representing the fraction of channels or dimensions removed from each layer. When applied symmetrically to both input and output channels of a convolutional layer, the parameter reduction is approximately
\[
\text{Parameter Reduction} = 1 - (1 - p)^2.
\]
For example, a pruning ratio of \(p=0.3\) (30\%) corresponds to around a 51\% reduction in parameters. 

\subsubsection{Knowledge Distillation}
\label{sec:kd_placeholder}

In the third stage of PDNet, we employ \emph{knowledge distillation} to recover the accuracy lost through structured pruning of the X3D-M backbone. 
Knowledge distillation is a model compression technique in deep learning where a large, pre-trained \textit{teacher} network transfers its knowledge to a smaller \textit{student} network, enabling the student to achieve competitive accuracy with fewer parameters and lower computational requirements \cite{Gou_2021}. The teacher generates softened probability outputs through a temperature-controlled softmax function:
\[
p_i = \frac{\exp(z_i / T)}{\sum_j \exp(z_j / T)},
\]
where \( z_i \) denotes the teacher logits and \( T > 1 \) is the temperature that smooths the distribution to reveal dark knowledge about inter-class relationships \cite{Gou_2021}. The student is trained to minimize a combination of the standard supervised loss and a distillation loss, which encourages it to mimic the teacher’s softened output distribution. Formally, the loss function is expressed as
\[
\mathcal{L} = \alpha \cdot \text{CE}(\mathbf{y}, \mathbf{\hat{y}}_s) + (1-\alpha) \cdot T^2 \cdot \text{KL}(p_t(T) \| p_s(T)),
\]
where \( \alpha \) balances the contribution of the ground-truth labels and teacher outputs, and \( p_t, p_s \) represents the teacher’s and student’s softened predictions respectively \cite{Gou_2021}.

Our  method is summarized in Algorithm \ref{algo}.


\begin{algorithm}[t]
\caption{Baseline Selection, Pruning, and Knowledge Distillation Fine-Tuning}
\DontPrintSemicolon
\SetKwInOut{Given}{Given}
\Given{candidate baseline models $\mathcal{F} = \{f_1, f_2, \ldots, f_k\}$, teacher model $f_{\text{teacher}}$, training set $X = \{(x_1, y_1), \ldots, (x_N, y_N)\}$}
\ForEach{$f_j \in \mathcal{F}$}{
    Train $f_j$ on $X$\;
}
\textbf{Select} model with best accuracy as $f_{\text{base}}$\;
initialize $f_{\text{pruned}} \leftarrow f_{\text{base}}$\;
initialize pruning ratio $r \leftarrow r_{\text{start}}$\;
\While{performance targets (size, latency, GFLOPS) not met \textbf{and} $r \le r_{\text{max}}$}{
    $f_{\text{pruned}} \leftarrow \text{prune}(f_{\text{base}}, r)$\;
    increase pruning ratio $r \leftarrow r + \Delta r$\;
}
\textbf{Update} $f_{\text{pruned}}$ by minimizing $L_{\text{KD}}(f_{\text{teacher}}(x), f_{\text{pruned}}(x), y)$ (aka knowledge distillation)\;
\Return{$f_{\text{pruned}}$ as final optimized model}
\label{algo}
\end{algorithm}

\subsection{Implementation Details}
The experiments were conducted using the PyTorch framework with the MECCANO dataset~\cite{meccanoDataset} as the primary benchmark. We trained 3D convolutional models such as 3D-MobileNet and X3D variants using cross-entropy loss and the SGD optimizer, with a base learning rate of 0.1, batch sizes of 12, 32, and 15 respectively, and a cosine annealing schedule over 120 epochs. Based on a trade-off analysis involving recognition accuracy, model size, and inference speed, the X3D-M model was selected as the optimal baseline. Model compression was performed using a structured pruning approach implemented via the \texttt{Torch-Pruning} library~\cite{torchPruningLibrary}, applying an L1-norm channel importance criterion at varying sparsity ratios. Each pruned model was then fine-tuned to recover part of the accuracy loss. To further mitigate the performance degradation introduced by pruning, we applied knowledge distillation using the pre-trained SlowFast model released by the MECCANO dataset authors as the teacher network and the pruned X3D-M model as the student. The student model was trained with a distillation temperature $T = 4$ and balancing coefficient $\alpha = 0.7$, using a batch size of 16, a base learning rate of 0.1, and the SGD optimizer for 180 epochs. The final optimized model achieved substantial reductions in computational cost while maintaining competitive accuracy suitable for real-time inference.

\section{\uppercase{Experiments}}

\subsection{Benchmark \& Metrics Details}
The models were benchmarked  on the MECCANO Dataset. MECCANO \cite{meccanoDataset} is a  egocentric dataset designed to model human-object interactions during mechanical assembly. It captures more than 20 hours of video from a first-person perspective, where users manipulate tools and parts to build a small mechanical system. The dataset features fine-grained activity labels and multiple annotators to ensure reliability. This dataset is unique for capturing 61 fine-grained action classes involved in complex assembly procedures and is characterized by a challenging long-tailed class distribution and high intra-class variability. 
\\
\\
Accuracy relied on standard classification metrics: Top-1 Test Accuracy ($\%$) and Top-5 Test Accuracy ($\%$). To accurately assess robustness across all activity types, particularly the rare ones, Mean Per-Class Accuracy was also calculated, addressing the inherent bias posed by the imbalanced class distribution.
\\
\\
Computational efficiency was evaluated using three metrics: model size (MB), inference time (ms), and computational cost (GFLOPs). The model size was determined by measuring the disk size of the saved \texttt{.pth} file. The inference time for a single video sample consisting of 16 frames was measured using the PyTorch profiler by averaging the forward-pass latency over 1000 iterations. The theoretical computational complexity, expressed in GFLOPs (billions of floating-point operations), represents the total number of operations required for a single forward pass of the model. All measurements were conducted on an AMD EPYC~7502P 32-core CPU.


\subsection{Results}

\paragraph{Baseline Model Selection}
Several compact 3D convolutional architectures, including 3D-MobileNet, X3D-L, and X3D-M, were evaluated using the PySlowFast framework against the high-capacity reference model SlowFast \cite{meccanoDataset}, which achieved the highest accuracy (45.16\% Top-1). Among the lightweight candidates, 3D-MobileNet achieved 27.31\% Top-1 and X3D-L achieved 34.50\% Top-1. Consequently, the X3D-M model was selected as the optimal baseline for subsequent optimization because it achieved the highest Top-1 accuracy (36.49\%) among compact models.

\begin{table}[!htbp]
\centering
\caption{Comparison of baseline models on the Meccano dataset against SlowFast Model}
\label{tab:baseline_comparison}
\renewcommand{\arraystretch}{1.4} 
\setlength{\tabcolsep}{6pt} 
\rowcolors{2}{gray!3}{white} 

\begin{tabular}{lcc}
\toprule
\rowcolor{gray!15}
\textbf{Model} & \textbf{Top-1 test (\%)} & \textbf{Top-5 test (\%)} \\
\midrule
\rowcolor{yellow!12}
SlowFast & 45.16 & 73.75\\
3D MobileNet & 27.31 & 50.51\\
X3D-L & 34.50 & 63.09\\
\rowcolor{green!12}
\textbf{X3D-M} & \textbf{36.49} & \textbf{65.64}\\
\bottomrule
\end{tabular}
\end{table}

\paragraph{Pruning Experiments}
To evaluate the effect of structured pruning on the baseline X3D-M model, we gradually increased the pruning ratio from no pruning up to 25\%. At each pruning ratio, the least important filters were removed to reduce model complexity while retaining as much representational capacity as possible. The resulting models were analyzed in terms of model size, latency, and computational cost (GFLOPs). The experiments showed that as the pruning ratio increased, model size and computational cost decreased, but this was accompanied by some degradation in accuracy. For example, pruning up to 25\% reduced the model size by approximately 41\% while maintaining a reasonable level of performance, highlighting the trade-off between efficiency and predictive capability.

\begin{table}[!htbp]
\centering
\caption{Pruning Results Summary}
\label{tab:pruning_results}
\renewcommand{\arraystretch}{1.5} 
\setlength{\tabcolsep}{6pt} 
\rowcolors{2}{gray!3}{white} 

\begin{tabular}{lcccc}
\toprule
\rowcolor{gray!15}
\textbf{Ratio (\%)} & \textbf{Size (MB)}& \textbf{Time (ms)} & \textbf{GFLOPs} \\
\midrule
\rowcolor{yellow!12}
0.0 & 21.70  & 107.72 & 6.20 \\
5.0 & 10.75 & 98.32 & 5.65 \\
10.0 & 9.71   & 83.75 & 5.14 \\
15.0 & 8.73  & 83.33 & 4.62 \\
20.0 & 7.78 & 81.17 & 4.18 \\
\rowcolor{green!12}
25.0 & 6.90 & 80.77 & 3.78 \\
\bottomrule
\end{tabular}
\end{table}

\begin{table*}[t!]
\centering
\caption{Comparison of the different variants of the X3D-M model. The pruning and fine-tuning approach achieves large efficiency gains while maintaining competitive accuracy.}
\label{tab:pruning_kd_results}
\renewcommand{\arraystretch}{1.4}
\setlength{\tabcolsep}{10pt}
\rowcolors{2}{gray!3}{white}

\begin{tabular}{lccccc}
\rowcolor{gray!15}
& \multicolumn{2}{c}{\textbf{Accuracy}} 
& \multicolumn{3}{c}{\textbf{Computational Efficiency}} \\
\cmidrule(lr){2-3} \cmidrule(lr){4-6}
\rowcolor{gray!15}
\textbf{Model Variant} & \textbf{Top-1 (\%)} & \textbf{Top-5 (\%)} 
& \textbf{Size (MB)} & \textbf{Time (ms)} & \textbf{GFLOPs} \\

\rowcolor{yellow!12}
Original X3D-M & 36.49 & 65.64 & 21.70 & 107.72 & 6.20 \\

Pruned Only & 25.54 & 58.64 & 6.90 & 80.77 & 3.78 \\

\rowcolor{green!12}
\textbf{PDNet (Ours)} & \textbf{39.46} & \textbf{68.72} 
& \textbf{6.90} & \textbf{80.77} & \textbf{3.78} \\

\bottomrule
\end{tabular}
\end{table*}


\paragraph{Main Results: Post-Pruning Accuracy Recovery through Knowledge Distillation}

The subsequent phase, focused on accuracy recovery, utilized Knowledge Distillation (KD) to fine-tune the highly compressed X3D-M student model using the high-capacity SlowFast teacher model. This combined Pruned + KD strategy was remarkably successful, yielding a final model that achieved a Top-1 accuracy of 39.46\% and a Top-5 accuracy of 68.72\%. Crucially, this optimization not only recovered the significant performance drop caused by pruning (which reduced Top-1 accuracy from 35.40\% to 25.54\%) but also enabled the student model to surpass the performance of the original uncompressed X3D-M baseline model by 4.06 percentage points (39.46\% vs.\ 35.40\%). However, the class-based accuracy analysis presented a more nuanced view: the overall mean class accuracy slightly declined from 12.01\% (original X3D-M) to 10.95\% (Pruned + KD model), a decrease of 1.06\%. This divergence between high Top-1 accuracy and low mean per-class accuracy is largely due to the highly imbalanced, long-tailed class distribution of the 61 fine-grained action classes present in the MECCANO dataset, where many rare actions are difficult to classify. Despite the slight drop in mean performance, KD significantly boosted accuracy for several key frequent classes, such as \texttt{check\_booklet} (improving by $+11.68\%$) and \texttt{tighten\_bolt\_with\_hands} ($+4.54\%$). Conversely, some visually or temporally complex fine-grained actions, such as \texttt{fit\_rim\_tire} ($-17.65\%$) and \texttt{put\_red\_4\_perforated\_junction\_bar} ($-20.00\%$), saw decreases, highlighting the limitations of the reduced model capacity for certain tail classes. Overall, the resulting Pruned + KD model maintained the efficiency of the pruned version (6.90~MB, 3.78~GFLOPs) while providing a substantial performance recovery, confirming the efficacy of this joint optimization approach for practical, real-time HAR deployment as shown in figure ~\ref{fig:performance_analysis}.


\section{CONCLUSIONS}

In this work, we proposed a resource-efficient optimization pipeline for 3D-CNNs tailored to real-time human activity recognition in industrial assembly settings. By selecting X3D-M as a compact baseline and applying structured pruning followed by knowledge distillation from a SlowFast teacher model, we achieved substantial reductions in model size and computational cost while improving Top-1 accuracy beyond the original model. The final compressed model maintains strong performance at only 3.78 GFLOPs and 6.90 MB, demonstrating the effectiveness of combining pruning and distillation for deployment in resource-constrained Industry~4.0 environments. While the current results are promising, further experiments are needed to fully explore the potential of this optimization pipeline. Future work could investigate alternative student-teacher combinations to identify more effective distillation pairings, evaluate different pruning strategies or sparsity levels, and assess the generalizability of the approach across other datasets or activity domains. Additionally, benchmarking the compressed models on a wider range of hardware platforms, including edge devices and industrial-grade embedded systems, will provide deeper insights into real-world performance and latency characteristics. These investigations will help refine the pipeline and ensure robust, efficient deployment in diverse industrial scenarios.

\section*{Acknowledgements}

The authors would like to acknowledge the use of AI-assisted tools, including Perplexity AI and ChatGPT, during the preparation of this manuscript. These tools were employed to support literature exploration, writing refinement, and editing.

\bibliographystyle{apacite}
\bibliography{example}

\end{document}